\documentclass[runningheads]{llncs}
\usepackage[T1]{fontenc}
\usepackage{amsmath} 
\usepackage{booktabs}
\usepackage{graphicx,verbatim}
\begin{document}
\title{Mask-Guided Attention Regulation for Anatomically Consistent Counterfactual CXR Synthesis}
%

\author{Zichun Zhang \and Weizhi Nie\thanks{Corresponding author.} \and Honglin Guo \and Yuting Su}
\institute{Tianjin University}
\maketitle              
\begin{abstract}

Counterfactual generation for chest X-rays (CXR) aims to simulate plausible pathological changes while preserving patient-specific anatomy. However, diffusion-based editing methods often suffer from structural drift, where stable anatomical semantics propagate globally through attention and distort non-target regions, and unstable pathology expression, since subtle and localized lesions induce weak and noisy conditioning signals. We present an inference-time attention regulation framework for reliable counterfactual CXR synthesis. An anatomy-aware attention regularization module gates self-attention and anatomy-token cross-attention with organ masks, confining structural interactions to anatomical ROIs and reducing unintended distortions. A pathology-guided module enhances pathology-token cross-attention within target lung regions during early denoising and performs lightweight latent corrections driven by an attention-concentration energy, enabling controllable lesion localization and extent. Extensive evaluations on CXR datasets show improved anatomical consistency and more precise, controllable pathological edits compared with standard diffusion editing, supporting localized counterfactual analysis and data augmentation for downstream tasks.

\keywords{Counterfactual generation  \and Diffusion model \and Attention regulation.}

\end{abstract}

\section{Introduction}
Counterfactual medical image generation aims to simulate plausible imaging outcomes under hypothetical changes in pathological conditions while preserving the underlying anatomical structure of the patient, thereby constructing clinically meaningful “what-if” imaging scenarios\cite{Cohen,WANG2025105747}. In chest X-ray (CXR) imaging, counterfactual generation allows localized pulmonary abnormalities to be introduced, removed, or progressively altered without disrupting stable anatomical components, including lung shape, rib structures, and cardiac contours\cite{ccalli2021deep}. By explicitly disentangling pathological variations from anatomical consistency, counterfactual medical image generation provides a powerful framework for improving model interpretability, supporting controlled data augmentation, and facilitating a more intuitive understanding of disease progression \cite{Choi_2021_10}.

With the rapid development of diffusion-based generative models\cite{ddpm}, a growing body of work has explored counterfactual medical image generation by leveraging image inpainting\cite{PIE}, instruction-driven editing\cite{pix2pix}\cite{biomedjourney}, or conditional synthesis\cite{emit}\cite{ctnet} frameworks. However, achieving stable and controllable editing often requires domain-specific retraining or learnable control branches\cite{ctnet}, adding recurring tuning and data-governance costs under cross-institution variability and continuously evolving data\cite{wang2021review}, which hinders deployment at scale.

Therefore, we aim to minimize additional training overhead by imposing constraints on the diffusion sampling process at inference time to improve the generality and controllability of our method. Under this setting, we still need to address the following key challenges:
\begin{itemize}
    \item \textbf{Structural instability.} 
    In diffusion models, global anatomical structures tend to stabilize early and dominate subsequent generation through self-attention\cite{com}. When pathology-related prompts are introduced, this global propagation can cause localized changes to spread to non-target regions\cite{x2x}, leading to unintended structural distortions and reduced anatomical consistency\cite{yang}.
    
    \item \textbf{Pathological expression instability.} 
    Pathological features in medical images are often subtle, spatially confined, and heterogeneous, resulting in weak attention responses during generation\cite{Cohen}. Consequently, such features may be suppressed or unintentionally diffused, leading to inaccurate localization or uncontrolled lesion extent\cite{devil}.
\end{itemize}

Motivated by these challenges, we propose an inference-time attention regulation framework for counterfactual medical image generation. Specifically, we first introduce an anatomy-aware attention regularization strategy that constrains self-attention responses\cite{trans} within valid anatomical regions, preventing structural semantics from excessively propagating into pathology-sensitive areas. Furthermore, to enable precise control over localized pathological changes, we design a pathology-guided attention regulation mechanism that enhances pathology-related cross-attention responses within target lung regions via a mask-derived spatial prior\cite{chen} and applies lightweight corrections to intermediate noise representations during denoising, allowing accurate lesion localization and extent control.

In summary, our contributions are threefold:
\begin{itemize}
    \item We propose an inference-time attention regulation framework for counterfactual medical image generation, which reduces repeated training and maintenance costs under cross-device and cross-domain shifts, improving generality and controllability.
    \item We jointly regularize anatomy-aware self-attention and pathology-guided cross-attention to preserve structure while enabling reliable, localized pathological edits.
    \item Extensive experiments on chest X-ray datasets demonstrate the effectiveness of the proposed framework in producing anatomically consistent and pathology-controllable counterfactual images.
\end{itemize}

\section{Method}
\begin{figure}[!t]
    \centering
    \includegraphics[width=1\linewidth]{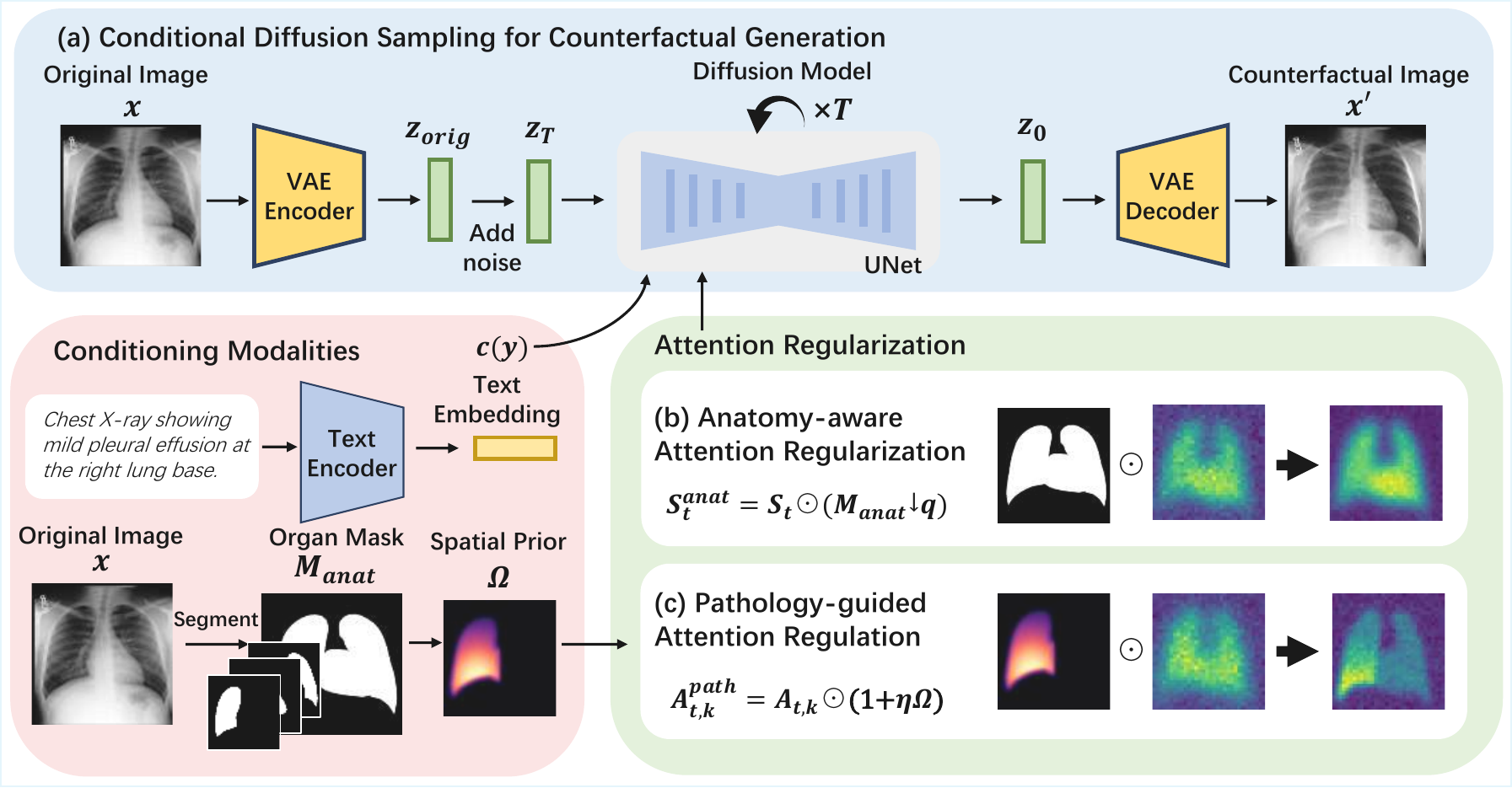}
    \caption{Overview of our inference-time attention regulation framework for counterfactual CXR generation. (a) The input image is encoded by VAE\cite{VAE}, noised, and denoised by a conditional diffusion model to generate a counterfactual image. (b) Anatomy-aware self-attention gating with $M_{\text{anat}}$ preserves structural consistency, while (c) pathology-guided cross-attention reweighting with a mask-derived prior $\Omega$ localizes pathological edits.}
    \label{fig:pipe}
\end{figure}
\subsection{Problem Definition}
Given an input medical image $x$, an organ mask $M_{\text{anat}}$, and a counterfactual condition $y$ describing the target pathological state, our goal is to generate a counterfactual image $x'$ that preserves anatomical consistency while realizing the desired pathological variation. We denote the counterfactual generator as
\begin{equation}
x' = \mathcal{G}(x, M_{\text{anat}}, c(y); \xi),
\label{eq:gen_map}
\end{equation}
where $\xi$ denotes the stochasticity.

As shown in Fig.\ref{fig:pipe} (a), in diffusion sampling, we first encode the input image $x$ into a latent $z_{\text{orig}}=\mathrm{Enc}(x)$ and obtain a noisy initialization by forward diffusion:
\begin{equation}
z_T = \sqrt{\bar{\alpha}_T}\, z_{\text{orig}} + \sqrt{1-\bar{\alpha}_T}\,\epsilon,\quad \epsilon\sim\mathcal{N}(0,I).
\label{eq:init_noisy}
\end{equation}
We then perform conditional iterative denoising:
\begin{equation}
z_{t-1} = F_{\theta}\!\left(z_t;\, x, M_{\text{anat}}, c(y), t\right), \quad t=T,\ldots,1,
\qquad x' = \mathrm{Dec}(z_0),
\label{eq:diff_process}
\end{equation}
where $F_{\theta}$ is the conditional denoising update implemented by a UNet\cite{unet} with parameters $\theta$, and $\mathrm{Dec}(\cdot)$ is the decoder.
In this work, we introduce lightweight inference-time regulation on the denoising updates in Eq.~\eqref{eq:diff_process}; the following sections describe our anatomy and pathology regulation strategies.

\subsection{Anatomy-aware Attention Regularization}
We consider a diffusion UNet where, at each denoising step $t$, intermediate latents are updated via attention blocks. Let $S_t$ denote the self-attention map capturing spatial interactions, and let $A_t$ denote the cross-attention map aligning conditioning tokens to spatial locations:
\begin{equation}
S_t=\mathrm{Softmax}\!\left(\frac{Q_{\text{self}}K_{\text{self}}^{\top}}{\sqrt{d}}\right),\qquad
A_t=\mathrm{Softmax}\!\left(\frac{Q_{\text{cross}}K^{\top}}{\sqrt{d}}\right),
\label{eq:attn_defs}
\end{equation}
where $Q_{\text{self}},K_{\text{self}}$ are projected from current features, and $K$ is projected from the condition embedding $c(y)$.
Given an organ mask $M_{\text{anat}}$, we denote by $(M_{\text{anat}}\downarrow q)$ its downsampled version to match the spatial resolution $q$ of the attention map at a specific UNet layer.

Stable anatomical structures tend to become dominant early in denoising and can be propagated through self-attention across the image\cite{com}. When pathological conditions are introduced, such global propagation may induce unintended structural drifts in non-target regions, degrading anatomical consistency. We therefore constrain anatomy-related attention responses to the corresponding anatomical regions using the available organ masks.

As shown in Fig.~\ref{fig:pipe} (b), we apply inference-time region gating to restrict attention-driven information flow by $M_{\text{anat}}$. For self-attention, we gate the attention response at each layer:
\begin{equation}
S_t^{\text{anat}}=S_t\odot (M_{\text{anat}}\downarrow q),
\label{eq:self_gate}
\end{equation}
which suppresses anatomy-driven interactions outside the anatomical ROI and reduces non-target structural perturbations during denoising.

\subsection{Pathology-guided Attention Regulation}
To realize counterfactual edits, we regulate token-wise cross-attention maps and apply a lightweight trajectory correction during inference as shown in Fig.~\ref{fig:pipe} (c). We interpret the token-wise cross-attention weight map $A_{t,k}$ as the spatial attribution of token $k$ over latent locations at denoising step $t$.
We construct a sample-specific spatial prior map $\Omega$ at the same resolution as $A_{t,k}$. 

Given the organ masks (e.g., left/right lung and heart), we deterministically select or combine an ROI mask $M_{\text{ROI}}$ using simple anatomy cues from the text $y$ (e.g., laterality and coarse lung regions), and obtain $\Omega$ by downsampling $M_{\text{ROI}}$ to the attention resolution followed by optional smoothing and normalization, i.e., $\Omega=\mathrm{Normalize}(\mathrm{Blur}(M_{\text{ROI}}\downarrow q))$.

Let $K_{\text{path}}$ denote the index set of pathology-related tokens, and let $A_{t,k}$ be the cross-attention map for token $k\in K_{\text{path}}$ at step $t$. During early denoising steps, we reweight token-wise cross-attention with a soft multiplier map $(1+\eta\Omega)$:
\begin{equation}
A_{t,k}^{\text{path}} = A_{t,k}\odot\Big(1+\eta\,\Omega\Big),\quad k\in K_{\text{path}},\ t<\mu T,
\label{eq:path_enhance}
\end{equation}
where $\eta$ controls the enhancement strength and $\mu$ specifies the early-step window. 

To mitigate attention diffusion and leakage to non-target areas during editing, we introduce a differentiable concentration metric that quantifies how well pathology-token attribution aligns with the target ROI, which further enables a lightweight inference-time correction. We define a region concentration score:
\begin{equation}
\mathrm{score}_{t,k}=\frac{\langle A_{t,k}^{\text{path}},\, \Omega\rangle}{\langle A_{t,k}^{\text{path}},\, \mathbf{1}\rangle},
\label{eq:score}
\end{equation}
and the pathology energy
\begin{equation}
L_{\text{path}}(t)=1-\frac{1}{|K_{\text{path}}|}\sum_{k\in K_{\text{path}}}\mathrm{score}_{t,k}.
\label{eq:lpath}
\end{equation}
We then apply a single inference-time correction to the intermediate latent within the same early-step window:
\begin{equation}
\hat{z}_t \leftarrow z_t - \alpha_t \nabla_{z_t} L_{\text{path}}(t),
\label{eq:z_update}
\end{equation}
where $\alpha_t$ is a step-size schedule. This lightweight update steers the denoising trajectory toward better ROI concentration, improving lesion localization and controlling the affected extent.

\subsection{Inference Procedure}
Given $(x, M_{\text{anat}}, y)$, we run conditional diffusion sampling for $t=T,\ldots,1$. 
At each denoising step, we first apply anatomy-aware self-attention regularization by gating self-attention with the downsampled organ mask in Eq.~\eqref{eq:self_gate}, thereby restricting attention-driven propagation to anatomically valid regions. 
We then perform pathology-guided cross-attention regulation for pathology tokens by reweighting token-wise cross-attention with the spatial prior $\Omega$ in Eq.~\eqref{eq:path_enhance}. 
During early steps ($t<\mu T$), we additionally apply a single latent correction update in Eq.~\eqref{eq:z_update} using the pathology energy in Eq.~\eqref{eq:lpath} before the denoising update. 
Finally, we decode the resulting $z_0$ to obtain the counterfactual image $x'$.

\section{Experiment}
\begin{figure}[!t]
    \centering
    \includegraphics[width=1\linewidth]{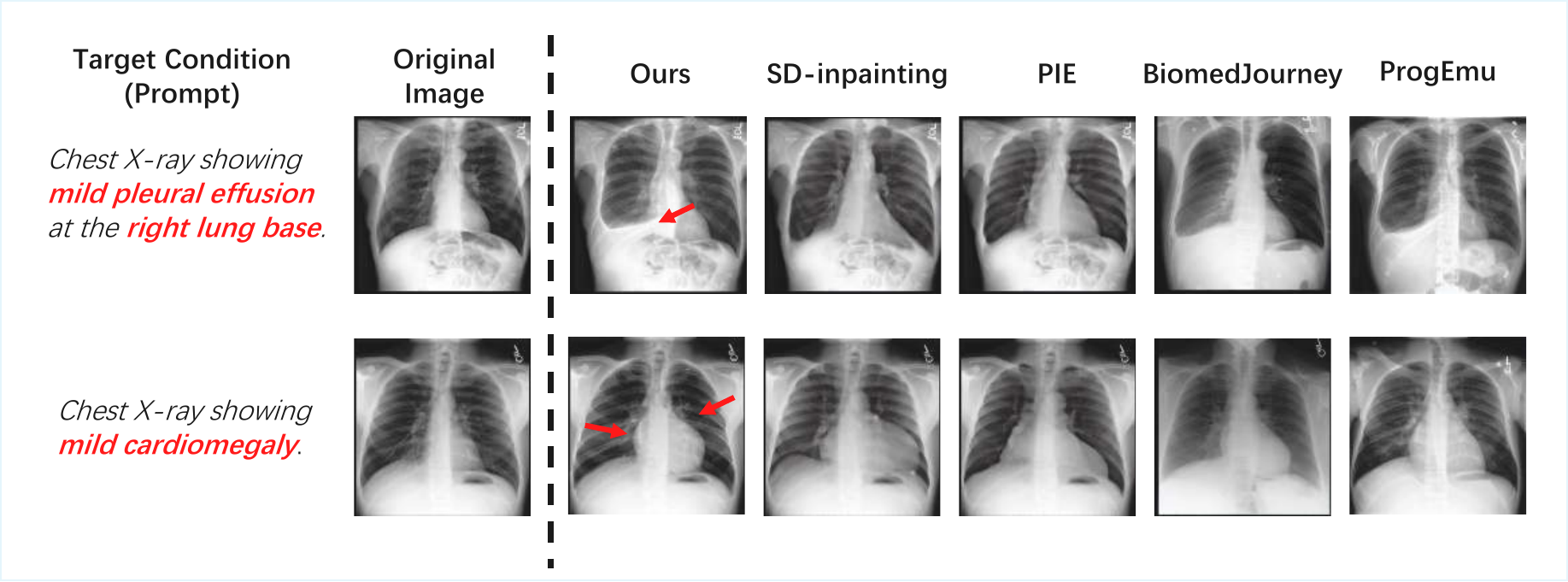}
    \caption{Qualitative comparison with state-of-the-art counterfactual CXR generation methods. Given the target prompt and the original image, we compare our results with SD-inpainting, PIE, BiomedJourney, and ProgEmu on representative cases.}
    \label{fig:sota}
\end{figure}
\textbf{Datasets.}We use frontal-view (PA) studies from MIMIC-CXR-JPG\cite{mimic} and ChexpertPlus\cite{cp}, resizing images to $512\times512$. Lung masks are obtained by HybridGNet\cite{hybridg}. We refine the paired report text using a GPT-based\cite{gpt} cleaning step that removes boilerplate and non-imaging content and compresses the remaining findings into concise prompts. Text conditions are truncated to fit the SD tokenizer limit (77 tokens).

\textbf{Implementation Details.} Our generator is built on Stable Diffusion v1.5\cite{ldm} and is fine-tuned on chest X-ray data for domain adaptation. The backbone is fine-tuned for 50 epochs using AdamW with a learning rate of $1{\times}10^{-4}$ and an effective batch size of 16; we use mixed precision (bf16) and gradient accumulation under GPU memory constraints. We perform DDIM sampling with $T{=}50$ steps and classifier-free guidance $s{=}7.5$. For attention regulation, we align the masks/prior to the attention resolution and use $q\in\{64,32,16\}$ for $512{\times}512$ inputs. Pathology-guided attention regulation is applied in an early window $t<\mu T$ with $\mu{=}0.4$ and enhancement strength $\eta{=}1.5$. Within the same window, latent correction uses one gradient step per denoising step with a linearly decayed step size $\alpha_t$, where $\alpha_{\max}{=}0.05$ at the start of the window.

\subsection{Comparison with State-of-the-Art}
\begin{table}[t]
\centering
\small
\setlength{\tabcolsep}{6pt}
\caption{Performance comparison with state-of-the-art methods on the counterfactual CXR generation task.}
\label{tab:sota_main}
\begin{tabular}{lcccc}
\toprule
Method & Conf$\uparrow$ & CLIP-I$\uparrow$ & LPIPS\cite{lps}$\downarrow$ & FID\cite{FID}$\downarrow$ \\
\midrule
SD-inpainting\cite{ldm}       & 0.662 & 0.827 & 0.23 & 37.6\\
Biomedjourney\cite{biomedjourney}     & 0.693& 0.868 & 0.19 & 30.8 \\
ProgEmu\cite{progemu}            & 0.701 & 0.866 &\textbf{ 0.17} & 28.4 \\
PIE\cite{PIE}                    & 0.691 & 0.842 & 0.18 & 29.7 \\
\midrule
Ours               & \textbf{0.709} & \textbf{0.870} & 0.18 & \textbf{29.0} \\
\bottomrule
\end{tabular}
\end{table}

We conduct both quantitative and qualitative comparisons with state-of-the-art methods for counterfactual CXR generation. As summarized in Table~\ref{tab:sota_main}, our method achieves the best overall performance when evaluated on a selected subset of the MIMIC-CXR-JPG dataset. For pathological accuracy, our approach provides more reliable and better-aligned target edits, indicating that the proposed inference-time regulation improves the intended pathological changes while maintaining global semantic consistency. The FID results suggest improved alignment with real-image distribution without sacrificing realism. For image quality, our results remain competitive in terms of perceptual realism and distribution-level fidelity, suggesting that the improved controllability does not come at the expense of visual plausibility.

Fig.~\ref{fig:sota} further shows representative visual examples. Compared with instruction-based editing methods, our generations exhibit stronger stability in background and non-target regions. Meanwhile, relative to prior editing/inpainting baselines, the pathological changes are more accurate and more tightly confined to relevant regions, demonstrating better preservation of irrelevant areas and validating the effectiveness of our framework.

\subsection{Ablation Study}
\begin{table}[t]
\centering
\small
\setlength{\tabcolsep}{6pt}
\caption{Ablation study of inference-time regulation modules on the counterfactual CXR generation task.}
\label{tab:ablation_est}
\begin{tabular}{lccc}
\toprule
Variant & Conf$\uparrow$ & CLIP-I$\uparrow$ & SSIM$\uparrow$ \\
\midrule
Full (Ours) & \textbf{0.71} & \textbf{0.87} & \textbf{0.80} \\
\midrule
w/o Anatomy self-attn regularization (Eq.~\ref{eq:self_gate}) & 0.69 & 0.85 & 0.76 \\
w/o Pathology cross-attn regulation (Eq.~\ref{eq:path_enhance}) & 0.66 & 0.82 & 0.80 \\
w/o Latent correction (Eq.~\ref{eq:z_update}) & 0.70 & 0.86 & 0.79 \\
\bottomrule
\end{tabular}
\end{table}

As shown in Table~\ref{tab:ablation_est}, we conduct an ablation study on the MIMIC-CXR-JPG dataset to evaluate pathological accuracy and the preservation of structural characteristics in counterfactual image generation.
Overall, the full configuration consistently yields the best results, indicating that neither anatomy stabilization nor pathology injection alone is sufficient for robust counterfactual generation. In particular, anatomy-aware self-attention gating primarily safeguards structural consistency and global appearance, while pathology-guided cross-attention regulation is the key factor for reliably expressing the intended pathological change. The latent correction step provides an additional layer of stability: although its impact is more modest, it consistently improves the final outcome by refining the denoising trajectory, leading to more dependable edits without over-amplifying the target signal.

\section{Conclusion}
In this work, we propose an inference-time attention regulation framework for counterfactual medical image generation to avoid recurring retraining and tuning costs. It addresses two practical challenges in diffusion-based medical editing: maintaining structural fidelity and producing accurate, localized pathological changes. By injecting organ-mask priors into attention modulation, our method suppresses unintended alterations in irrelevant regions while strengthening pathology-focused attention within target areas to better realize stage-specific edits. Extensive experiments demonstrate improved anatomical consistency and semantic alignment over strong diffusion editing baselines, highlighting its potential for disease progression modeling and clinically meaningful data augmentation.

%
%
%
\bibliographystyle{splncs04}
\bibliography{mybibliography}

@inproceedings{Cohen,
  title={Gifsplanation via Latent Shift: A Simple Autoencoder Approach to Counterfactual Generation for Chest X-rays},
  booktitle={International Conference on Medical Imaging with Deep Learning (MIDL)},
  author={Cohen, Joseph Paul and Brooks, Rupert and Zucker, Evan and Pareek, Anuj and Lungren, Matthew P. and Chaudhari, Akshay},
  year={2021},
  pages={74--104}
}

@article{Choi_2021_10,
  title={Ilvr: Conditioning method for denoising diffusion probabilistic models},
  author={Choi, Jooyoung and Kim, Sungwon and Jeong, Yonghyun and Gwon, Youngjune and Yoon, Sungroh},
  journal={arXiv preprint arXiv:2108.02938},
  year={2021}
}

@article{ddpm,
  title={Denoising diffusion probabilistic models},
  author={Ho, Jonathan and Jain, Ajay and Abbeel, Pieter},
  journal={Advances in neural information processing systems},
  volume={33},
  pages={6840--6851},
  year={2020}
}

@article{emit, title={EMIT-Diff: Enhancing Medical Image Segmentation via Text-Guided Diffusion Model}, DOI={10.48550/arxiv.2310.12868}, journal={arXiv.org}, author={Zhang, Zheyu and Yao, Lanhong and Wang, Bin and Jha, Debesh and Keles, Elif and A. Medetalibeyoğlu and Bagci, Ulas}, year={2023}, language={en} }

@misc{PIE,
      title={PIE: Simulating Disease Progression via Progressive Image Editing}, 
      author={Kaizhao Liang and Xu Cao and Kuei-Da Liao and Tianren Gao and Wenqian Ye and Zhengyu Chen and Jianguo Cao and Tejas Nama and Jimeng Sun},
      year={2023},
      eprint={2309.11745},
      archivePrefix={arXiv},
      primaryClass={eess.IV},
      url={https://arxiv.org/abs/2309.11745}, }

@article{biomedjourney,
  title={Biomedjourney: Counterfactual biomedical image generation by instruction-learning from multimodal patient journeys},
  author={Gu, Yu and Yang, Jianwei and Usuyama, Naoto and Li, Chunyuan and Zhang, Sheng and Lungren, Matthew P and Gao, Jianfeng and Poon, Hoifung},
  journal={arXiv preprint arXiv:2310.10765},
  year={2023}
}

@article{com,
  title={CompCraft: Foreground-Driven Image Synthesis With Customized Layouts},
  author={Guo, Honglin and Chen, Ruidong and Nie, Weizhi and Wang, Lanjun and Liu, Anan},
  journal={IEEE Transactions on Circuits and Systems for Video Technology},
  year={2025},
  publisher={IEEE}
}

@article{trans,
  title={Attention is all you need},
  author={Vaswani, Ashish and Shazeer, Noam and Parmar, Niki and Uszkoreit, Jakob and Jones, Llion and Gomez, Aidan N and Kaiser, {\L}ukasz and Polosukhin, Illia},
  journal={Advances in neural information processing systems},
  volume={30},
  year={2017}
}

@inproceedings{chen,
  title={Training-free layout control with cross-attention guidance},
  author={Chen, Minghao and Laina, Iro and Vedaldi, Andrea},
  booktitle={Proceedings of the IEEE/CVF winter conference on applications of computer vision},
  pages={5343--5353},
  year={2024}
}

@inproceedings{pix2pix,
  title={Instructpix2pix: Learning to follow image editing instructions},
  author={Brooks, Tim and Holynski, Aleksander and Efros, Alexei A},
  booktitle={Proceedings of the IEEE/CVF conference on computer vision and pattern recognition},
  pages={18392--18402},
  year={2023}
}

@inproceedings{ctnet,
  title={Adding conditional control to text-to-image diffusion models},
  author={Zhang, Lvmin and Rao, Anyi and Agrawala, Maneesh},
  booktitle={Proceedings of the IEEE/CVF international conference on computer vision},
  pages={3836--3847},
  year={2023}
}

@inproceedings{unet,
  title={U-net: Convolutional networks for biomedical image segmentation},
  author={Ronneberger, Olaf and Fischer, Philipp and Brox, Thomas},
  booktitle={International Conference on Medical image computing and computer-assisted intervention},
  pages={234--241},
  year={2015},
  organization={Springer}
}

@misc{cp,
      title={CheXpert Plus: Augmenting a Large Chest X-ray Dataset with Text Radiology Reports, Patient Demographics and Additional Image Formats}, 
      author={Pierre Chambon and Jean-Benoit Delbrouck and Thomas Sounack and Shih-Cheng Huang and Zhihong Chen and Maya Varma and Steven QH Truong and Chu The Chuong and Curtis P. Langlotz},
      year={2024},
      eprint={2405.19538},
      archivePrefix={arXiv},
      primaryClass={cs.CL},
      url={https://arxiv.org/abs/2405.19538}, 
}

@article{mimic,
  title={MIMIC-CXR, a de-identified publicly available database of chest radiographs with free-text reports},
  author={Johnson, Alistair EW and Pollard, Tom J and Berkowitz, Seth J and Greenbaum, Nathaniel R and Lungren, Matthew P and Deng, Chih-ying and Mark, Roger G and Horng, Steven},
  journal={Scientific data},
  volume={6},
  number={1},
  pages={317},
  year={2019},
  publisher={Nature Publishing Group UK London}
}

@article{hybridg,
  title={Improving anatomical plausibility in medical image segmentation via hybrid graph neural networks: applications to chest x-ray analysis},
  author={Gaggion, Nicol{\'a}s and Mansilla, Lucas and Mosquera, Candelaria and Milone, Diego H and Ferrante, Enzo},
  journal={IEEE Transactions on Medical Imaging},
  volume={42},
  number={2},
  pages={546--556},
  year={2022},
  publisher={IEEE}
}

@misc{progemu,
      title={Towards Interpretable Counterfactual Generation via Multimodal Autoregression}, 
      author={Chenglong Ma and Yuanfeng Ji and Jin Ye and Lu Zhang and Ying Chen and Tianbin Li and Mingjie Li and Junjun He and Hongming Shan},
      year={2025},
      eprint={2503.23149},
      archivePrefix={arXiv},
      primaryClass={eess.IV},
      url={https://arxiv.org/abs/2503.23149}, 
}

@inproceedings{ldm,
  title={High-resolution image synthesis with latent diffusion models},
  author={Rombach, Robin and Blattmann, Andreas and Lorenz, Dominik and Esser, Patrick and Ommer, Bj{\"o}rn},
  booktitle={Proceedings of the IEEE/CVF conference on computer vision and pattern recognition},
  pages={10684--10695},
  year={2022}
}

@misc{VAE,
  title={Auto-encoding variational bayes},
  author={Kingma, Diederik P and Welling, Max and others},
  year={2013},
  publisher={Banff, Canada}
}

@article{FID,
  title={Gans trained by a two time-scale update rule converge to a local nash equilibrium},
  author={Heusel, Martin and Ramsauer, Hubert and Unterthiner, Thomas and Nessler, Bernhard and Hochreiter, Sepp},
  journal={Advances in neural information processing systems},
  volume={30},
  year={2017}
}

@inproceedings{lps,
  title={The unreasonable effectiveness of deep features as a perceptual metric},
  author={Zhang, Richard and Isola, Phillip and Efros, Alexei A and Shechtman, Eli and Wang, Oliver},
  booktitle={Proceedings of the IEEE conference on computer vision and pattern recognition},
  pages={586--595},
  year={2018}
}

@article{yang,
  title={Dynamic prompt learning: Addressing cross-attention leakage for text-based image editing},
  author={Yang, Fei and Yang, Shiqi and Butt, Muhammad Atif and van de Weijer, Joost and others},
  journal={Advances in Neural Information Processing Systems},
  volume={36},
  pages={26291--26303},
  year={2023}
}

@inproceedings{x2x,
  title={InstructX2X: An Interpretable Local Editing Model for Counterfactual Medical Image Generation},
  author={Min, Hyungi and You, Taeseung and Lee, Hangyeul and Cho, Yeongjae and Cho, Sungzoon},
  booktitle={International Conference on Medical Image Computing and Computer-Assisted Intervention},
  pages={279--288},
  year={2025},
  organization={Springer}
}

@article{devil,
  title={The Devil is in Attention Sharing: Improving Complex Non-rigid Image Editing Faithfulness via Attention Synergy},
  author={Chen, Zhuo and Wei, Fanyue and Xu, Runze and Li, Jingjing and Duan, Lixin and Yao, Angela and Li, Wen},
  journal={arXiv preprint arXiv:2512.14423},
  year={2025}
}

@article{wang2021review,
  title={A review on medical imaging synthesis using deep learning and its clinical applications},
  author={Wang, Tonghe and Lei, Yang and Fu, Yabo and Wynne, Jacob F and Curran, Walter J and Liu, Tian and Yang, Xiaofeng},
  journal={Journal of applied clinical medical physics},
  volume={22},
  number={1},
  pages={11--36},
  year={2021},
  publisher={Wiley Online Library}
}

@article{ccalli2021deep,
  title={Deep learning for chest X-ray analysis: A survey},
  author={{\c{C}}all{\i}, Erdi and Sogancioglu, Ecem and Van Ginneken, Bram and Van Leeuwen, Kicky G and Murphy, Keelin},
  journal={Medical image analysis},
  volume={72},
  pages={102125},
  year={2021},
  publisher={Elsevier}
}

@article{gpt,
  title={Gpt-4 technical report},
  author={Achiam, Josh and Adler, Steven and Agarwal, Sandhini and Ahmad, Lama and Akkaya, Ilge and Aleman, Florencia Leoni and Almeida, Diogo and Altenschmidt, Janko and Altman, Sam and Anadkat, Shyamal and others},
  journal={arXiv preprint arXiv:2303.08774},
  year={2023}
}

@article{WANG2025105747,
title = {MBT-Polyp: A new Multi-Branch Memory-augmented Transformer for polyp segmentation},
journal = {Image and Vision Computing},
volume = {163},
pages = {105747},
year = {2025},
issn = {0262-8856},
doi = {https://doi.org/10.1016/j.imavis.2025.105747},
url = {https://www.sciencedirect.com/science/article/pii/S026288562500335X},
author = {Tao Wang and Weijie Wang and Fausto Giunchiglia and Fengzhi Zhao and Ye Zhang and Duo Yu and Guixia Liu}
}

\end{document}